\documentclass[11pt,a4paper]{article}
\usepackage[hyperref]{acl2021}
\usepackage{times}
\usepackage{latexsym}

\usepackage{microtype}
\usepackage{multirow}
\usepackage{amsfonts}
\usepackage{graphicx}
\usepackage{amsmath}
\usepackage{subfigure}
\usepackage{amsmath}
\usepackage{color}

\aclfinalcopy 
\def\aclpaperid{2172} 

\title{DialogueCRN: Contextual Reasoning Networks for Emotion Recognition in Conversations}

\author{Dou Hu$^{1}$, Lingwei Wei$^{2,3}$, Xiaoyong Huai$^{1}$ \\
$^1$ National Computer System Engineering Research Institute of China \\ 
$^2$ Institute of Information Engineering, Chinese Academy of Sciences \\
$^3$ School of Cyber Security, University of Chinese Academy of Sciences \\
  \texttt{\{hudou18, weilingwei18\}@mails.ucas.edu.cn} { } \\
  \texttt{huaixy@sina.com} \\ }
  
\date{}

\begin{document}
\maketitle
\begin{abstract}
Emotion Recognition in Conversations (ERC) has gained increasing attention for developing empathetic machines. Recently, many approaches have been devoted to perceiving conversational context by deep learning models. However, these approaches are insufficient in understanding the context due to lacking the ability to extract and integrate emotional clues. In this work, we propose novel Contextual Reasoning Networks (DialogueCRN) to fully understand the conversational context from a cognitive perspective. Inspired by the Cognitive Theory of Emotion, we design multi-turn reasoning modules to extract and integrate emotional clues. The reasoning module iteratively performs an intuitive retrieving process and a conscious reasoning process, which imitates human unique cognitive thinking. Extensive experiments on three public benchmark datasets demonstrate the effectiveness and superiority of the proposed model. 

\end{abstract}

\section{Introduction}
Emotion recognition in conversation (ERC) aims to detect emotions expressed by the speakers in each utterance of the conversation. 
The task is an important topic for developing empathetic machines
\cite{DBLP:journals/coling/ZhouGLS20} in a variety of areas including social opinion mining \cite{DBLP:conf/ic3/KumarDD15}, intelligent assistant \cite{DBLP:conf/ph/KonigFMH16}, health care \cite{DBLP:conf/riiforum/PujolMM19}, and so on.

A conversation often contains contextual clues \cite{DBLP:conf/acl/PoriaHMNCM19} that trigger the current utterance's emotion, such as the cause or situation.
Recent context-based works \cite{DBLP:conf/acl/PoriaCHMZM17,DBLP:conf/naacl/HazarikaPZCMZ18,DBLP:conf/aaai/MajumderPHMGC19} on ERC have been devoted to perceiving situation-level or speaker-level context by deep learning models.
However, these methods are insufficient in understanding the context that usually contains rich emotional clues.
We argue they mainly suffer from the following challenges.
1) \textbf{The extraction of emotional clues}. Most approaches \cite{DBLP:conf/emnlp/HazarikaPMCZ18,DBLP:conf/naacl/HazarikaPZCMZ18,DBLP:conf/aaai/JiaoLK20} generally retrieve the relevant context from a static memory, which limits the ability to capture richer emotional clues.
2) \textbf{The integration of emotional clues}. 
Many works \cite{DBLP:conf/aaai/MajumderPHMGC19,DBLP:conf/emnlp/GhosalMPCG19,DBLP:conf/coling/LuZWTCQ20} usually use the attention mechanism to integrate encoded emotional clues, ignoring their intrinsic semantic order.  It would lose logical relationships between clues, making it difficult to capture key factors that trigger emotions.  

The \textit{Cognitive Theory of Emotion} \cite{article1962,scherer2001appraisal} suggests that cognitive factors are potently determined for the formation of emotional states.
These cognitive factors can be captured by iteratively performing the intuitive retrieving process and conscious reasoning process in our brains \cite{evans1984heuristic,evans2003two,evans2008dual,sloman1996empirical}. 
Motivated by them, this paper attempts to model both critical processes to reason emotional clues and sufficiently understand the conversational context.
By following the mechanism of \textit{working memory} \cite{baddeley1992working} in the cognitive phase, we can iteratively perform both cognitive processes to guide the extraction and integration of emotional clues, which imitates human unique cognitive thinking.

In this work, we propose novel Contextual Reasoning Networks (DialogueCRN) to recognize the utterance's emotion by sufficiently understanding the conversational context.
The model introduces a cognitive phase to extract and integrate emotional clues from the context retrieved by the perceive phase.
Firstly, in the perceptive phase, we leverage Long Short-Term Memory (LSTM) \cite{DBLP:journals/neco/HochreiterS97} networks to capture situation-level and speaker-level context.
Based on the above context, global memories can be obtained to storage different contextual information.
Secondly, in the cognitive phase, we design multi-turn reasoning modules to iteratively extract and integrate the emotional clues.
The reasoning module performs two processes, {\it i.e.}, an intuitive retrieving process and a conscious reasoning process. 
The former utilizes the attention mechanism to match relevant contextual clues by retrieving static global memories, which imitates the intuitive retrieving process.
The latter adopts LSTM networks to learn intrinsic logical order and integrate contextual clues by retaining and updating dynamic working memory, which imitates the conscious reasoning process. It is slower but with human-unique rationality \cite{baddeley1992working}.
Finally, according to the above contextual clues at situation-level and speaker-level, an emotion classifier is used to predict the emotion label of the utterance.


To evaluate the performance of the proposed model, we conduct extensive experiments on three public benchmark datasets, {\it i.e.,} {\it IEMOCAP}, {\it SEMAINE} and {\it MELD} datasets. 
Results consistently demonstrate that our proposed model significantly outperforms comparison methods. 
Moreover, understanding emotional clues from a cognitive perspective can boost the performance of emotion recognition.

The main contributions of this work are summarized as follows:
\begin{itemize}
  \item We propose novel Contextual Reasoning Networks (DialogueCRN) to fully understand the conversational context from a cognitive perspective. 
  To the best of our knowledge, this is the first attempt to explore cognitive factors for emotion recognition in conversations. 
  \item We design multi-turn reasoning modules to extract and integrate emotional clues by iteratively performing the intuitive retrieving process and conscious reasoning process, which imitates human unique cognitive thinking.
  \item We conduct extensive experiments on three public benchmark datasets. The results consistently demonstrate the effectiveness and superiority of the proposed model\footnote{ The source code is available at { \url{https://github.com/zerohd4869/DialogueCRN}}}. 

\end{itemize}

\section{Methodology}

\subsection{Problem Statement}
Formally, let $U=[u_1, u_2, ..., u_N]$ be a conversation, where $N$ is the number of utterances.
And there are $M$ speakers/parties $p_1, p_2, ..., p_M\ (M \geq 2)$.
Each utterance $u_i$ is spoken by the speaker $p_{\phi(u_i)}$, where $\phi$ maps the index of the utterance into that of the corresponding speaker.
Moreover, for each $\lambda \in [1,M]$, we define $U_{{\lambda}}$ to represent the set of utterances spoken by the speaker $p_{\lambda}$, {\it i.e.}, $U_{{\lambda}} = \{ u_i\ |\ u_i \in U \text{ and }  u_i \text{ spoken by }   p_{\lambda}, \text{ } \forall i \in [1, N]   \}.$

The task of emotion recognition in conversations (ERC) aims to predict the emotion label $y_i$ for each utterance $u_i$ from the pre-defined emotions $\mathcal{Y}$.

\begin{figure*}[t]
  \centering
  \includegraphics[width=0.99\linewidth]{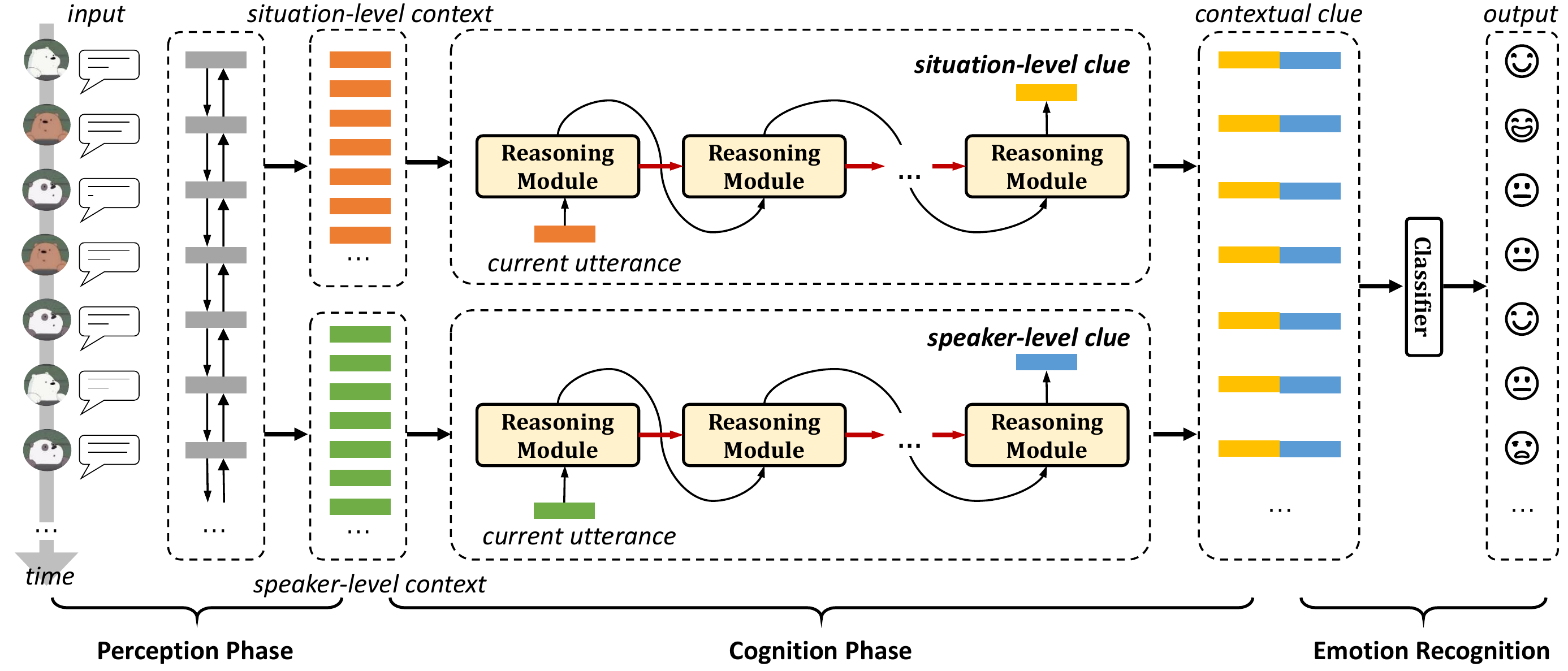}
  \caption{The architecture of the proposed model DialogueCRN. }
  \label{fig:overall}
\end{figure*}

\subsection{Textual Features}
Convolutional neural networks (CNNs) \cite{DBLP:conf/emnlp/Kim14} are capable of capturing n-grams information from an utterance. 
Following previous works \cite{DBLP:conf/naacl/HazarikaPZCMZ18,DBLP:conf/aaai/MajumderPHMGC19,DBLP:conf/emnlp/GhosalMPCG19},
we leverage a CNN layer with max-pooling to exact context-free textual features from the transcript of each utterance.
Concretely, the input is the 300 dimensional pre-trained 840B GloVe vectors  \cite{DBLP:conf/emnlp/PenningtonSM14}.
We employ three filters of size $3,4$ and $5$ with $50$ feature maps each.
These feature maps are further processed by max-pooling and ReLU activation  \cite{DBLP:conf/icml/NairH10}.
Then, these activation features are concatenated and finally projected onto a dense layer with dimension $d_u=100$, whose output forms the representation of an utterance.
We denote $\{ \textbf{u}_i\}_{i=1}^N, \textbf{u}_i \in \mathbb{R}^{d_u}$ as the representation for $N$ utterances.

\subsection{Model}

Then, we propose Contextual Reasoning Networks (DialogueCRN) for emotion recognition in conversations. 
DialogueCRN is comprised of three integral components, {\it i.e.,} the perception phase (Section \ref{sec:per}), the cognition phase (Section \ref{sec:cog}), and an emotion classifier (Section \ref{sec:emo}). 
The overall architecture is illustrated in Figure~\ref{fig:overall}.

\subsubsection{Perception Phase} \label{sec:per}

In the perceptive phase, based on the input textual features, we first generate the representation of conversational context at situation-level and speaker-level.
Then, global memories are obtained to storage different contextual information.

\paragraph{Conversational Context Representation.}
Long Short-Term Memory (LSTM) \cite{DBLP:journals/neco/HochreiterS97} introduces the gating mechanism into recurrent neural networks to capture long-term dependencies from the input sequences.  
In this part, two bi-directional LSTM networks are leveraged to capture situation-level and speaker-level context dependencies, respectively.

For learning the context representation at the situation level, we apply a bi-directional LSTM network to capture sequential dependencies between adjacent utterances in a conversational situation.
The input is each utterance's textual features $\textbf{u}_i \in \mathbb{R}^{d_u}$. 
The situation-level context representation $\textbf{c}^s_{i} \in \mathbb{R}^{2d_u}$ can be computed as:
\begin{equation}
  \textbf{c}^s_i, \textbf{h}^s_{i} = {\overleftrightarrow{LSTM}}^s(\textbf{u}_{i}, \textbf{h}^s_{i-1}),
\end{equation} 
where 
$\textbf{h}^s_{i} \in \mathbb{R}^{d_u}$ is the $i$-th hidden state of the situation-level LSTM.

For learning the context representation at the speaker level, we also employ another bi-directional LSTM network to capture self-dependencies between adjacent utterances of the same speaker. 
Given textual features $\textbf{u}_i$ of each utterance, the speaker-level context representation $\textbf{c}^v_i \in \mathbb{R}^{2d_u}$ is computed as:
\begin{equation}
  \resizebox{0.89\linewidth}{!}{$
  \textbf{c}^v_i, \textbf{h}_{\lambda,j}^v = \overleftrightarrow{LSTM}^v (\textbf{u}_{i}, \textbf{h}_{\lambda,j-1}^v),   j \in [1,|U_{\lambda}|], 
  $}
\end{equation} 
where $\lambda = \phi(u_i)$.  
$U_\lambda$ refers to all utterances of the speaker $p_\lambda$. $\textbf{h}^v_{\lambda, j} \in \mathbb{R}^{d_u} $ is the $j$-th hidden state of speaker-level LSTM for the speaker $p_\lambda$.

\paragraph{Global Memory Representation.}
Based on the above conversational context representation, global memories can be obtained to storage different contextual information via a linear layer.
That is, global memory representation of situation-level context $\textbf{G}^s = [\textbf{g}^s_1, \textbf{g}^s_2, ...,\textbf{g}^s_N ] $ and that of speaker-level context $\textbf{G}^v=[\textbf{g}^v_1, \textbf{g}^v_2, ..., \textbf{g}^v_N]$ can be computed as:
\begin{align}
  \textbf{g}_i^s &= \textbf{W}^s_g \textbf{c}^s_i + \textbf{b}^s_g, \\ 
  \textbf{g}_i^v &= \textbf{W}^v_g \textbf{c}^v_i + \textbf{b}_g^v, 
\end{align}
where $\textbf{W}^s_g,\textbf{W}^v_g \in \mathbb{R}^{{2d_u} \times 2{d_u} }$, $\textbf{b}^s_g, \textbf{b}^v_g \in \mathbb{R}^{ 2{d_u}}$ are learnable parameters.

\subsubsection{Cognition Phase} \label{sec:cog}

\begin{figure}[t]
  \centering
  \includegraphics[width=0.76\linewidth]{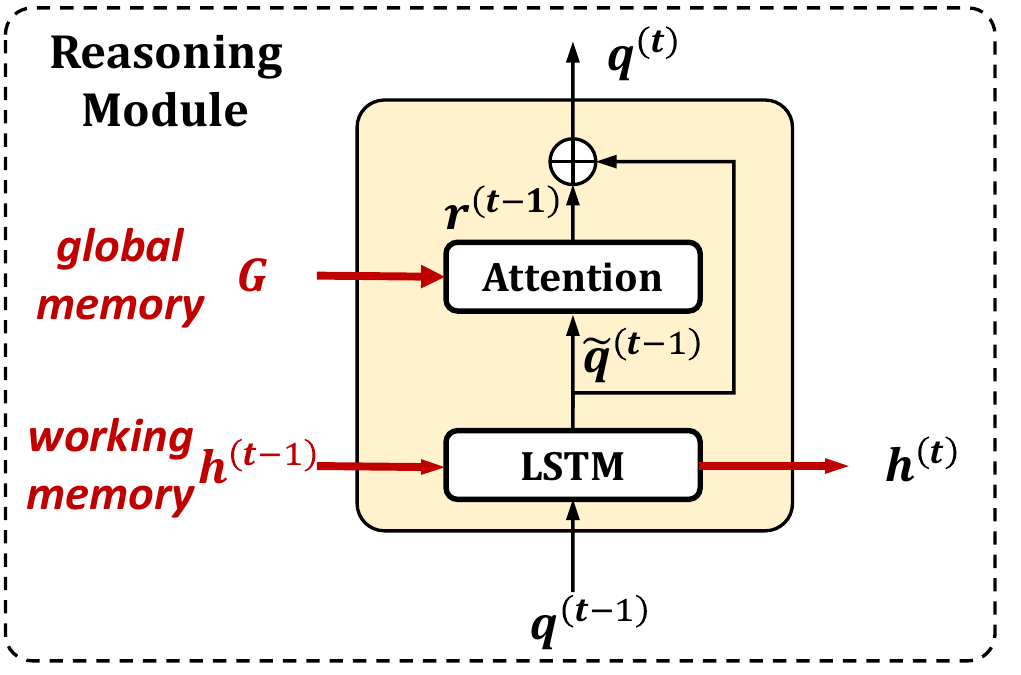}
  \caption{The detailed structure of reasoning module. }
  \label{fig:reason}
\end{figure}

Inspired by the \textit{Cognitive Theory of Emotion} \cite{article1962,scherer2001appraisal}, cognitive factors are potently determined for the formation of emotional states.
Therefore, in the cognitive phase, we design multi-turn reasoning modules to iteratively extract and integrate the emotional clues.
The architecture of a reasoning module is depicted in Figure \ref{fig:reason}.

The reasoning module performs two processes, the intuitive retrieving process, and the conscious reasoning process. 
In the $t$-th turn, for the \textbf{reasoning process}, we adopt the LSTM network to learn intrinsic logical order and integrate contextual clues in the working memory, which is slower but with human-unique rationality \cite{baddeley1992working}.
That is, 
\begin{equation}
  \tilde{\textbf{q}}_{i}^{(t-1)}, \textbf{h}_{i}^{(t)} = \overrightarrow{LSTM}({\textbf{q}}^{(t-1)}_{i}, \textbf{h}^{(t-1)}_i), \label{eq:lstm}
\end{equation} 
where $\tilde{\textbf{q}}_{i}^{(t-1)} \in \mathbb{R}^{2d_u}$ is the output vector. $\textbf{q}_i^{(t)} \in \mathbb{R}^{4d_u} $ is initialized by the context representation $\textbf{c}_i$ of the current utterance,
\textit{i.e.}, $\textbf{q}^{(0)}_i = \textbf{W}_q \textbf{c}_i + \textbf{b}_q$,
where $\textbf{W}_q \in \mathbb{R}^{4d_u \times 2d_u}$ and $\textbf{b}_q \in \mathbb{R}^{4d_u}$ are learnable parameters.
$\textbf{h}_{i}^{(t)} \in \mathbb{R}^{2d_u}$ refers to the working memory, which
can not only storage and update the previous memory $\textbf{h}_{i}^{(t-1)}$, but also guide the extraction of clues in the next turn.
During sequential flowing of the working memory, we can learn implicit logical order among clues, which resembles the conscious thinking process of humans.
$\textbf{h}_{i}^{(t)}$ is initialized with zero.
$t$ is the index that indicates how many ``processing steps" are being carried to compute the final state.

For the \textbf{retrieving process}, we utilize an attention mechanism to match relevant contextual clues from the global memory.
The detailed calculations are as follows:
\begin{align}
  \textbf{e}^{(t-1)}_{ij} &= f(\textbf{g}_j, \tilde{\textbf{q}}_{i}^{(t-1)}), \label{eq:b} \\
  {\alpha}_{ij}^{(t-1)} &= \frac {\exp(\textbf{e}_{ij}^{(t-1)})} {\sum_{j=1}^{N} \exp({\textbf{e}_{ij}^{(t-1)}}) },  \label{eq:c} \\ 
  \textbf{r}^{(t-1)}_i &= \sum_{j=1}^N \alpha_{ij}^{(t-1)} \textbf{g}_j,  \label{eq:d} 
\end{align}
where $f$ is a function that computes a single scalar from $\textbf{g}_j$ and $\tilde{\textbf{q}}_{i}^{(t-1)}$  ({\it e.g.}, a dot product). 

Then, we concatenate the output of reasoning process $\tilde{\textbf{q}}^{(t-1)}_i$ with the resulting attention readout $\textbf{r}^{(t-1)}_i$ to form the next-turn query ${\textbf{q}}_{i}^{(t)}$. 
That is,
\begin{equation}
      {\textbf{q}}^{(t)}_i = [\tilde{\textbf{q}}^{(t-1)}_i; \textbf{r}^{(t-1)}_i]. \label{eq:e}
\end{equation}
The query ${\textbf{q}}_{i}^{(t)}$ will be updated under the guidance of working memory $\textbf{h}_{i}^{(t)}$, and more contextual clues can be retrieved from the global memory.

To sum up, given context representation $\textbf{c}_i$ of the utterance $u_i$, global memory representation $\textbf{G}$, and the number of turns $T$, the whole cognitive phase (Eq.\ref{eq:lstm}-\ref{eq:e}) can be denoted as, $\textbf{q}_i = Cognition(\textbf{c}_i, \textbf{G}; T)$.
In this work, we design two individual cognition phases to explore contextual clues at situation-level and speaker-level, respectively. The outputs are defined as:
\begin{align}
\textbf{q}_i^s &= Cognition^s(\textbf{c}^s_i, \textbf{G}^s; T^s), \\ 
\textbf{q}_i^v &= Cognition^v(\textbf{c}^v_i, \textbf{G}^v; T^v),
\end{align}
where $T^s$ and $T^v$ are the number of turns in situation-level and speaker-level cognitive phases, respectively. 

Based on the above output vectors, the final representation $\textbf{o}$ can be defined as a concatenation of both vectors, \textit{i.e.},
\begin{equation}
  \textbf{o}_i = [\textbf{q}^s_i;\textbf{q}^v_i]. 
\end{equation}

\subsubsection{Emotion Classifier} \label{sec:emo}
Finally, according to the above contextual clues, an emotion classifier is used to predict the emotion label of the utterance.
\begin{equation}
    \hat{\textbf{y}}_i = softmax(\mathbf{W}_o \textbf{o}_i + \mathbf{b}_o),    
\end{equation} 
where $\mathbf{W}_o \in \mathbb{R}^{8d_u \times |\mathcal{Y}|}$ and $\mathbf{b}_o \in \mathbb{R}^{|\mathcal{Y}|}$ are trainable parameters.
$|\mathcal{Y}|$ is the number of emotion labels.

Cross entropy loss is used to train the model. The loss function is defined as:
\begin{equation}
    \mathcal{L} = - \frac{1}{\sum_{l=1}^L \tau(l)} \sum_{i=1}^{L} \sum^{\tau(i)}_{k=1} {\textbf{y}}^l_{i,k} log (\hat{\textbf{y}}^l_{i,k}), 
  \end{equation}    
where $L$ is the total number of conversations/samples in the training set.
$\tau(i)$ is the number of utterances in the sample $i$. 
$\textbf{y}^l_{i,k}$ and $\hat{\textbf{y}}^l_{i,k}$ denote the one-hot vector and probability vector for emotion class $k$ of utterance $i$ of sample $l$, respectively.

\section{Experimental Setups}

\subsection{Datasets}
We evaluate our proposed model on following benchmark datasets, {\it IEMOCAP} \cite{DBLP:journals/lre/BussoBLKMKCLN08}, {\it SEMAINE} \cite{DBLP:journals/taffco/McKeownVCPS12}, and {\it MELD} \cite{DBLP:conf/acl/PoriaHMNCM19} datasets. 
The statistics are reported in Table \ref{tab:datasets}.
The above datasets are multimodal datasets with textual, visual, and acoustic features. In this paper, we focus on emotion recognition in textual conversations. Multimodal emotion recognition in conversations is left as future work.

    \textbf{IEMOCAP}\footnote{\url{https://sail.usc.edu/iemocap/}}: The dataset \cite{DBLP:journals/lre/BussoBLKMKCLN08} contains videos of two-way conversations of ten unique speakers, where only the first eight speakers from session one to four belong to the training set. The utterances are annotated with one of six emotion labels, namely {\it happy}, {\it sad}, {\it neutral}, {\it angry}, {\it excited}, and {\it frustrated}.  
    Following previous works \cite{DBLP:conf/emnlp/HazarikaPMCZ18,DBLP:conf/emnlp/GhosalMPCG19,DBLP:conf/aaai/JiaoLK20}, 
    the validation set is extracted from the randomly shuffled training set with the ratio of 80:20 since no pre-defined train/val split is provided in the {\it IEMOCAP} dataset. 

    \textbf{SEMAINE}\footnote{\url{https://semaine-db.eu}}: 
    The dataset \cite{DBLP:journals/taffco/McKeownVCPS12} is a video database of human-agent interactions. 
    It is available at AVEC 2012's {\it fully continuous sub-challenge} \cite{DBLP:conf/icmi/SchullerVCP12} that requires predictions of four continuous affective attributes: {\it Arousal}, {\it Expectancy}, {\it Power}, and
    {\it Valence}.
    The gold annotations are available for every 0:2 seconds in each video \cite{DBLP:conf/icmi/NicolleRBPC12}. Following \cite{DBLP:conf/emnlp/HazarikaPMCZ18,DBLP:conf/emnlp/GhosalMPCG19}, the attributes are averaged over the span of an utterance to obtain utterance-level annotations.
    We utilize the standard both training and testing splits provided in the sub-challenge. 

    \textbf{MELD}\footnote{\url{https://github.com/SenticNet/MELD}}: 
     Multimodal Emotion Lines Dataset (MELD) \cite{DBLP:conf/acl/PoriaHMNCM19}, a extension of the EmotionLines \cite{DBLP:conf/lrec/HsuCKHK18},
     is collected from TV-series Friends containing more than 1400 multi-party conversations and 13000 utterances. Each utterance is annotated with one of seven emotion labels (\textit{i.e.}, {\it happy}/{\it joy}, {\it anger}, {\it fear},  {\it disgust}, {\it sadness}, {\it surprise}, and {\it neutral}).
     We use the pre-defined train/val split provided in the \textit{MELD} dataset.

\begin{table}[t]
\centering  
\resizebox{\linewidth}{!}{$
\begin{tabular}{ccccccccc}
    \hline
    \multirow{2}{*}{\textbf{Dataset}} &  \multicolumn{3}{ c }{\textbf{\# Dialogues}}  & \multicolumn{3}{c}{\textbf{\# Utterances}} & 
    \multirow{1}{*}{\textbf{Avg.} } &\multirow{2}{*}{\textbf{\# Classes}} \\
    \cline{2-7} 
        & \textbf{\textit{train}} & \textbf{\textit{val}} & \textbf{\textit{test}}
        & \textbf{\textit{train}} & \textbf{\textit{val}}   & \textbf{\textit{test}} 
        & \multirow{1}{*}{\textbf{Length} } &  \\
    \hline
    IEMOCAP & \multicolumn{2}{c}{120} & 31 & \multicolumn{2}{c}{5,810} & 1,623 & 50 & 6 \\  
    SEMAINE & \multicolumn{2}{c}{63} & 32 & \multicolumn{2}{c}{4,368} & 1,430 & 72 & 4$^*$ \\  
    MELD & 1,039 & 114 & 280 & 9,989 &  1,109 & 2,610 & 10 & 7  \\ \hline
    \multicolumn{9}{l}{$^*$ refers to the number of real valued attributes.} 
    \end{tabular}
    $}
    \caption{The statistics of three datasets.}
    \label{tab:datasets}
\end{table}


\subsection{Comparisons Methods}
We compare the proposed model against the following baseline methods.
    \textbf{TextCNN} \cite{DBLP:conf/emnlp/Kim14} is a convolutional neural network trained on context-independent utterances.
    \textbf{Memnet} \cite{DBLP:conf/nips/SukhbaatarSWF15} is an end-to-end memory network
    and update memories in a multi-hop fashion.    
    \textbf{bc-LSTM+Att} \cite{DBLP:conf/acl/PoriaCHMZM17} adopts a bidirectional LSTM network to capture the contextual content from the surrounding utterances. Additionally, an attention mechanism is adopted to re-weight features and provide a more informative output.
    \textbf{CMN} \cite{DBLP:conf/naacl/HazarikaPZCMZ18} encodes conversational context from dialogue history by two distinct GRUs for two speakers. 
    \textbf{ICON} \cite{DBLP:conf/emnlp/HazarikaPMCZ18} extends CMN by connecting outputs of individual speaker GRUs using another GRU for perceiving inter-speaker modeling.
    \textbf{DialogueRNN} \cite{DBLP:conf/aaai/MajumderPHMGC19} is a recurrent network that consists of two GRUs to track speaker states and context during the conversation.
    \textbf{DialogueGCN} \cite{DBLP:conf/emnlp/GhosalMPCG19} 
    a graph-based model where nodes represent utterances and edges represent the dependency between the speakers of the utterances.

\subsection{Evaluation Metrics}
Following previous works  \cite{DBLP:conf/emnlp/HazarikaPMCZ18,DBLP:conf/aaai/JiaoLK20}, for {\it IEMOCAP} and {\it MELD} datasets, we choose the {\bf accuracy score} ({\it Acc}.) to measure the overall performance. We also report the {\bf Weighted-average F1 score} ({\it Weighted}-$F1$) and {\bf Macro-averaged F1 score} ({\it Macro}-$F1$) to evaluate the model performance on both majority and minority classes, respectively. 
For the {\it SEMAINE} dataset, we report {\bf Mean Absolute Error} ({\it MAE}) for each attribute. The lower {\it MAE}, the better the detection performance.


\subsection{Implementation Details}
We use the validation set to tune hyperparameters. 
In the perceptive phase, we employ two-layer bi-directional LSTM on {\it IEMOCAP} and {\it SEMAINE} datasets and single-layer bi-directional LSTM on the {\it MELD} dataset.   
In the cognitive phase, single-layer LSTM is used on all datasets.
The batch size is set to 32.
We adopt Adam \cite{DBLP:journals/corr/KingmaB14} as the optimizer with an initial learning rate of \{0.0001, 0.001, 0.001\} and L2 weight decay of \{0.0002, 0.0005, 0.0005\} for {\it IEMOCAP}, {\it SEMAINE}, {\it MELD} datasets, respectively. 
The dropout rate is set to $0.2$.  
We train all models for a maximum of $100$ epochs and stop training if the validation loss does not decrease for 20 consecutive epochs.

For results of DialogueGCN and DialogueRNN, we implement them according to the public code\footnote{\url{https://github.com/declare-lab/conv-emotion}} provided by \citet{DBLP:conf/aaai/MajumderPHMGC19,DBLP:conf/emnlp/GhosalMPCG19} under the same environment. 

\begin{table}[t]
  \centering
      \resizebox{0.96\linewidth}{!}{$
  \begin{tabular}{lccc}
  \hline
    \multicolumn{4}{ c }{\textbf{IEMOCAP}} 
  \\ 
  \hline
  \multirow{1}{*}{\textbf{Methods}} & {\textit{Acc.}}& {\textit{Weighted-F$1$}}  
   & {\textit{Macro-F$1$}}  \\
  \hline
 TextCNN & 49.35 & 49.21 & 48.13 \\ 
  Memnet & 55.70 & 53.10 & 55.40  \\
  bc-LSTM+Att & 56.32 & 56.19 & 54.84  \\ 
  CMN & 56.56 & 56.13 & 54.30  \\
  ICON & 59.09 & 58.54 & 56.52  \\ 
  DialogueRNN & 63.03 & 62.50 & 60.66  \\ 
  DialogueGCN& 64.02  & 63.65 & 63.43   \\ 
  \hline 
  DialogueCRN &  {\bf 66.05} & {\bf 66.20} & {\bf 66.38}  \\ 
  {\bf Improve}  & {\bf 3.2\%} & {\bf 4.0\%} & {\bf 4.7\%} \\
  \hline
  \end{tabular}    
  $}
  \caption{Experimental results on the {\it IEMOCAP} dataset. } \label{tab:iemocap}
\end{table}

\begin{table}[t]
\centering
    \resizebox{0.99\linewidth}{!}{$
\begin{tabular}{lcccc}
\hline
  \multicolumn{5}{c}{\textbf{SEMAINE}} 
\\ 
\hline
\multirow{2}{*}{\textbf{Methods}} &  \multicolumn{4}{c}{\textit{MAE}}  \\
 \cline{2-5}
 & {\textit{Valence}}& {\textit{Arousal}} & {\textit{Expectancy}} & {\textit{Power}} \\
\hline
TextCNN & 0.545 & 0.542 & 0.605 & 8.71 \\ 
Memnet & 0.202 & 0.211 & 0.216 & 8.97 \\
bc-LSTM+Att & 0.189 & 0.213 & 0.190 & 8.67 \\
CMN & 0.192 & 0.213 & 0.195 & 8.74 \\ 
ICON & 0.180 & 0.190 & 0.180 & 8.45 \\
DialogueRNN & 0.175 & 0.171 & 0.181 & 8.66 \\
DialogueGCN& 0.176 & 0.210 & 0.193 & 8.65 \\
\hline 
DialogueCRN &  {\bf 0.173} & {\bf 0.152} & {\bf 0.175} & {\bf 8.20} \\ 
{\bf Improve}  & {\bf 1.1\%} & {\bf 11.1\%} & {\bf 2.8\%} & {\bf 2.9\%} \\
\hline
\end{tabular}
$}
\caption{Experimental results on the {\it SEMAINE} dataset. } \label{tab:semaine}
\end{table}

\begin{table}[t]
\centering
    \resizebox{0.96\linewidth}{!}{$
\begin{tabular}{lccc}
\hline
  \multicolumn{4}{ c }{\textbf{MELD}} 
\\ 
\hline
\multirow{1}{*}{\textbf{Methods}} & {\textit{Acc.}}& {\textit{Weighted-F$1$}}  
 & {\textit{Macro-F$1$}}  \\
\hline
TextCNN & 59.69 & 56.83 & 33.80 \\ 
bc-LSTM+Att & 57.50 & 55.90 & 34.84  \\ 
CMN & - & 54.50 &  -  \\
ICON & - & 54.60 &  - \\ 
DialogueRNN & 59.54 & 56.39 & 32.93  \\ 
DialogueGCN& 59.46 & 56.77 & 34.05   \\ 
\hline 
DialogueCRN &  {\bf 60.73} & {\bf 58.39} & {\bf 35.51}  \\ 
{\bf Improve}  & {\bf 2.0\%} & {\bf 2.9\%} & {\bf 1.9\%} \\
\hline
\end{tabular}
$}
\caption{Experimental results on the {\it MELD} dataset. } \label{tab:meld}
\end{table}

\section{Results and Analysis}
\subsection{Experimental Results}


Table \ref{tab:iemocap}, \ref{tab:semaine} and \ref{tab:meld} show the comparison results for emotion recognition in textual conversations.
DialogueCRN consistently achieves better performance than the comparison methods on all datasets,
while also being statistically significant under the paired $t$-test (p$<$0.05).

\paragraph{\textit{IEMOCAP} and \textit{SEMAINE}.} Both {\it IEMOCAP} and {\it SEMAINE} datasets have long conversation lengths and the average length is not less than 50.
The fact implies that the two datasets contain richer contextual information. 
\textbf{TextCNN} ignoring conversational context obtains the worst performance.
\textbf{Memnet} and \textbf{bc-LSTM+Att} perceive the situation-level context of the current utterance. \textbf{CMN} perceives the speaker-level context.
Thereby, \textbf{Memnet}, \textbf{bc-LSTM+Att} and \textbf{CMN} slightly outperforms \textbf{TextCNN}.
\textbf{ICON}, \textbf{DialogueRNN}, and \textbf{DialogueGCN} consider both situation-level and speaker-level context to model the perceptive phase of context.
They achieve better performance than the above methods.
Compared with baseline methods, \textbf{DialogueCRN} can extract and integrate rich emotional clues by exploring cognitive factors. 
Accordingly, our model obtains more effective performance.
That is, as shown in Table \ref{tab:iemocap} and \ref{tab:semaine}, for the \textit{IEMOCAP} dataset, \textbf{DialogueCRN} gains 3.2\%, 4.0\%, 4.7\% relative improvements over the previous best baselines in terms of \textit{Acc.}, \textit{Weighted-F}$1$, and \textit{Macro-F}$1$, respectively.
For the \textit{SEMAINE} dataset, \textbf{DialogueCRN} achieves a large margin of 11.1\% MAE for the \textit{Arousal} attribute.

\paragraph{\textit{MELD}.}
From Table~\ref{tab:datasets}, the number of speakers of each conversation in the {\it MELD} dataset is large (up to 9), and the average length of conversations is 10.
The shorter conversation length of the {\it MELD} dataset indicates it contains less contextual information.
From the result in Table~\ref{tab:meld}, interestingly, \textbf{TextCNN} ignoring conversational context achieves better results than most baselines.
It indicates that it is difficult to learn useful features from perceiving a limited and missing context.
Besides, \textbf{DialogueGCN} leverages graph structure to perceive the interaction of multiple speakers, which is sufficient to perceive the speaker-level context. Thereby, the performance is slightly improved. 
Compared with baselines, \textbf{DialogueCRN} enables to perform sequential thinking of context and understand emotional clues from a cognitive perspective. 
Therefore, it achieves the best recognition results, \textit{e.g.},  2.9\% improvements on \textit{Weighted-F}$1$.

\subsection{Ablation Study}

\begin{table*}[t]
  \centering
      \resizebox{0.98\linewidth}{!}{$
  \begin{tabular}{cccc|ccc|cccc}
  \hline
   \multicolumn{2}{ c }{\textit{Cognition}}  & \multicolumn{2}{ c| }{\textit{Perception}}   &  \multicolumn{3}{ c| }{\textbf{IEMOCAP}}
    &  \multicolumn{4}{ c }{\textbf{SEMAINE}}
     \\ 
  \cline{1-11}
  {\textit{Situation}}& {\textit{Speaker}} 
  & {\textit{Situation}}& {\textit{Speaker}} 
  & \multirow{2}{*}{\textit{Acc.}} & \multirow{2}{*}{\textit{Weighted-F$1$}} & \multirow{2}{*}{\textit{Macro-F$1$}}  &  \multicolumn{4}{ c }{\textit{MAE}}  \\ \cline{8-11}
  {\textit{Context}}& {\textit{Context}} 
  & {\textit{Context}}& {\textit{Context}} 
  & & & &  {\it Valence} &	{\it Arousal} &	{\it Expectancy} &	{\it Power}
  \\
  \hline
    $\checkmark$ & $\checkmark$ & $\checkmark$ & $\checkmark$ & {\bf 66.05} & {\bf 66.20} & {\bf 66.38} 
  & {\bf 0.173} &	{\bf 0.152} &		{\bf 0.175} &		{\bf 8.201} \\ \hline 
   $\checkmark$ & $\times$ & $\checkmark$ & $\checkmark$ &  64.26 & 64.43 & 62.86
  & 0.173& 0.162 & 0.181 & 8.253 
   \\ 
   $\times$ & $\checkmark$ & $\checkmark$ & $\checkmark$ & 63.28 & 63.8 & 63.31  
  & 0.174 & 0.165 & 0.179 & 8.201
   \\   
  \hline
   $\times$ & $\times$ & $\checkmark$ & $\checkmark$ & 63.22 & 63.37 & 62.08 
  & 0.177 & 0.171 & 0.180 & 8.237
  \\ 
   $\times$ & $\times$ & $\times$ & $\checkmark$ &63.50 & 63.68 & 62.40 
  & 0.192 & 0.213 & 0.195 & 8.740
  \\  
   $\times$ & $\times$ & $\checkmark$ & $\times$ & 60.07 & 60.14 & 59.58 
  & 0.194 & 0.212 & 0.201 & 8.900
  \\ 
   $\times$ & $\times$ & $\times$ & $\times$ & 49.35 & 49.21 & 48.13 
  & 0.545 & 0.542 & 0.605 & 8.710
  \\    
  \hline
  \end{tabular}
  $}
  \caption{Experimental results of ablation studies on {\it IEMOCAP} and {\it SEMAINE} datasets. } \label{tab:ablation}
  \end{table*}

To better understand the contribution of different modules in DialogueCRN to the performance, we conduct several ablation studies on both \textit{IEMOCAP} and \textit{SEMAINE} datasets.
Different modules that model the situation-level and speaker-level context in both perceptive and cognitive phases are removed separately.
The results are shown in Table~\ref{tab:ablation}. 
When cognition and perception modules are removed successively, the performance is greatly declined.
It indicates the importance of both the perception and cognition phases for ERC.

\textbf{Effect of Cognitive Phase.}
When only removing cognition phase, as shown in the third block of Table~\ref{tab:ablation},
the performance on the \textit{IEMOCAP} dataset decreases 4.3\%, 4.3\% and 6.5\% in terms of \textit{Acc.}, \textit{Weighted-F}$1$, and  \textit{Macro-F}$1$, respectively.
And on the \textit{SEMAINE} dataset, the {\it MAE} scores of {\it Valence}, {\it Arousal}, and {\it Expectancy} attributes are increased by 2.3\%, 12.5\% and 2.9\%, respectively. 
These results indicate the efficacy of the cognitive phase, which can reason based on the perceived contextual information consciously and sequentially.
Besides, if removing the cognitive phase for either speaker-level or situation-level context, as shown in the second block, the results decreased on both datasets.
The fact reflects both situational factors and speaker factors are critical in the cognitive phase.

\textbf{Effect of Perceptive Phase.}
As shown in the last row, when removing the perception module, the performance is dropped sharply.
The inferior results reveal the necessity of the perceptive phase to unconsciously match relevant context based on the current utterance.

\textbf{Effect of Different Context.}
When removing either situation-level or speaker-level context in both cognitive and perceptive phases, respectively, the performance has a certain degree of decline.
The phenomenon shows both situation-level and speaker-level context play an effective role in the perceptive and cognitive phases. 
Besides, the margin of dropped performance is different on both datasets.
This suggests speaker-level context plays a greater role in the perception phase while more complex situation-level context works well in the cognitive phase.
The explanation is that it is limited to learn informative features from context by intuitive matching perception, but conscious cognitive reasoning can boost better understanding.

\begin{figure}[t]
  \centering
  \includegraphics[width=0.76\linewidth]{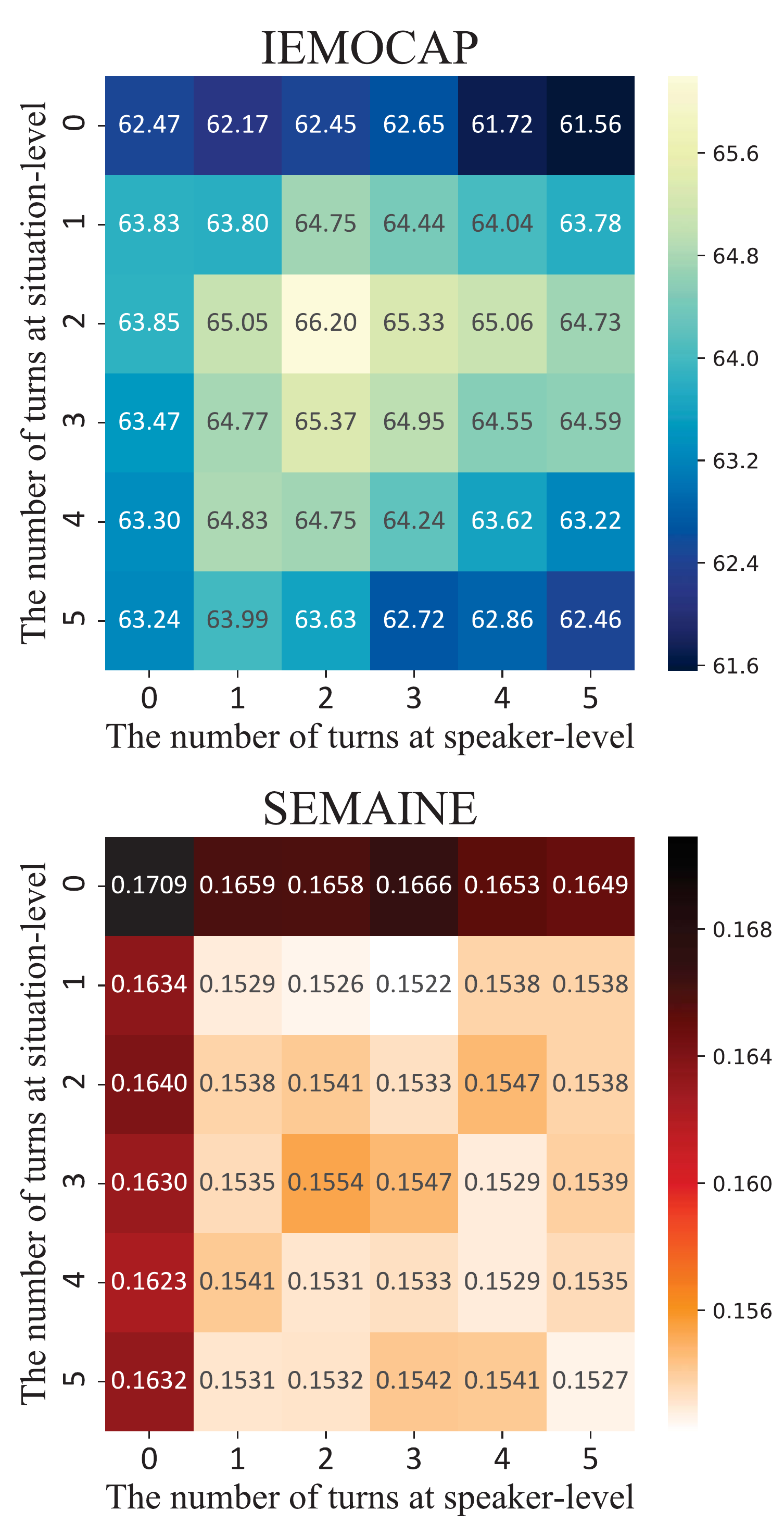} 
  \caption{Results against the number of turns. We report the \textit{Weighted-F}$1$ score on the {\it IEMOCAP} dataset and {\it MAE} of \textit{Arousal} attribute on the \textit{SEMAINE} dataset. The lighter the color, the better the performance.
  }
  \label{fig:param}
\end{figure}

\begin{figure}[t]
  \centering
  \includegraphics[width=\linewidth]{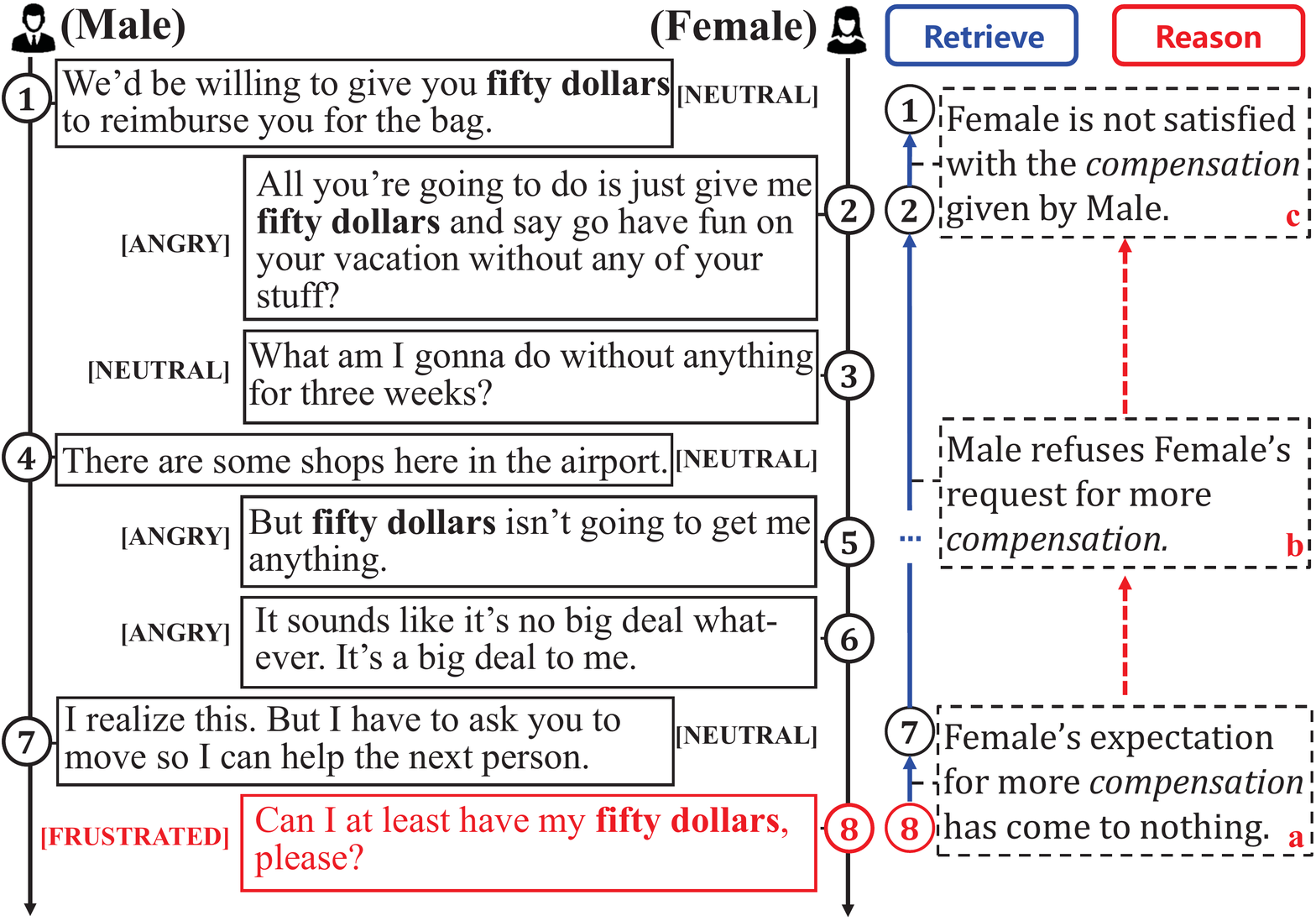}
  \caption{The case study. }
  \label{fig:case}
\end{figure}

\subsection{Parameter Analysis}
We investigate how our model performs w.r.t the number of turns in the cognitive phase.
From Figure~\ref{fig:param}, the best $\{ T^s, T^v \}$ is $\{2,2\}$ and $\{1,3\}$ on {\it IEMOCAP} and {\it SEMAINE} datasets, which obtain 66.20\% \textit{Weighted-F}$1$ and 0.1522 \textit{MAE} of \textit{Arousal} attribute, respectively.
Note that the {\it SEMAINE} dataset needs more turns for the speaker-level cognitive phase. It implies speaker-level contextual clues may be more vital in arousal emotion, especially empathetic clues that require complex reasoning.

Besides, if we solely consider either situation-level or speaker-level context in the cognitive phase, results on the two datasets are significantly improved within a certain number of turns.
The fact indicates the effectiveness of using multi-turn reasoning modules to understand contextual clues.


\subsection{Case Study}
Figure~\ref{fig:case} shows a conversation sampled from the \textit{IEMOCAP} dataset.
The goal is to predict the emotion label of \textit{utterance} 8.
Methods such as DialogueRNN and DialogueGCN lack the ability to consciously understand emotional clues, {\it e.g.}, the cause of the emotion (failed expectation). They are easy to mistakenly identify the emotion as {\it angry} or {\it neutral}. 

Our model DialogueCRN can understand the conversational context from a cognitive perspective. 
In the cognitive phase, the following two processes are performed iteratively:
the intuitive retrieving process of {\it 8-7-2-1} (\textcolor[RGB]{5,3,128}{blue} arrows) and the conscious reasoning process of  {\it a-b-c} (\textcolor{red}{red} arrows), to extract and integrate emotional clues.
We can obtain that {\it utterance} 8 implied that more compensation expected by {\it female} was not achieved.
The failed compensation leads to more negative of his emotion and thus correctly identified as {\it depression}. 

\section{Related Work}

\subsection{Emotion Recognition}
Emotion recognition (ER) has been drawing increasing attention to natural language processing (NLP) and artificial intelligence (AI).
Existing works generally regard the ER task as a classification task based on context-free blocks of data, such as individual reviews or documents.  
They can roughly divided into two parts, \textit{i.e.}, feature-engineering based \cite{DBLP:conf/interspeech/DevillersV06},  and deep-learning based methods 
\cite{DBLP:conf/coling/TangQFL16,wei2020hierarchical}.

\subsection{Emotion Recognition in Conversations}
Recently, the task of Emotion Recognition in Conversations (ERC) has received attention from researchers.
Different traditional emotion recognition, both situation-level and speaker-level context plays a significant role in identifying the emotion of an utterance in conversations \cite{DBLP:conf/coling/LiJLZL20}.
The neglect of them would lead to quite limited performance \cite{DBLP:conf/emnlp/BerteroSWWCF16}.
Existing works generally capture contextual characteristics for the ERC task by deep learning methods, which can be divided into {\it sequence-based} and {\it graph-based} methods.

\textbf{Sequence-based Methods.}
Many works capture contextual information in utterance sequences. 
\citet{DBLP:conf/acl/PoriaCHMZM17} employed LSTM \cite{DBLP:journals/neco/HochreiterS97} to capture conversational context features.
\citet{DBLP:conf/emnlp/HazarikaPMCZ18,DBLP:conf/naacl/HazarikaPZCMZ18} used end-to-end memory networks \cite{DBLP:conf/nips/SukhbaatarSWF15} to capture contextual features that distinguish different speakers.
\citet{DBLP:conf/emnlp/ZhongWM19,DBLP:conf/coling/LiJLZL20} utilized the transformer \cite{DBLP:conf/nips/VaswaniSPUJGKP17} to capture richer contextual features based on the attention mechanism.
\citet{DBLP:conf/aaai/MajumderPHMGC19} introduced a speaker state and global state for each conversation based on GRUs \cite{DBLP:conf/emnlp/ChoMGBBSB14}.
Moreover, 
\citet{DBLP:conf/emnlp/JiaoLK20} introduced a conversation completion task to learn from unsupervised conversation data.
\citet{DBLP:conf/aaai/JiaoLK20} proposed a hierarchical memory network for real-time emotion recognition without future context.
\citet{DBLP:conf/sigdial/WangZMWX20} modeled ERC as sequence tagging to learn the emotional consistency.
\citet{DBLP:conf/coling/LuZWTCQ20} proposed an iterative emotion interaction network to explicitly model the emotion interaction.

\textbf{Graph-based Methods.}
Some works \cite{DBLP:conf/ijcai/ZhangWSLZZ19,DBLP:conf/emnlp/GhosalMPCG19,DBLP:conf/emnlp/IshiwatariYMG20,DBLP:conf/interspeech/LianTL0YL20a} model the conversational context by designing a specific graphical structure.
They utilize graph neural networks \cite{DBLP:conf/iclr/KipfW17,DBLP:journals/corr/abs-1710-10903} to capture multiple dependencies in the conversation, which have achieved appreciable performance.

Different from previous works, inspired by the \textit{Cognitive Theory of Emotion} \cite{article1962,scherer2001appraisal}, this paper makes the first attempt to explore cognitive factors for emotion recognition in conversations. To sufficiently understand the conversational context, we propose a novel DialogueCRN to extract and then integrate rich emotional clues in a cognitive manner. 



\section{Conclusion}
This paper has investigated cognitive factors for the task of emotion recognition in conversations (ERC). We propose novel contextual reasoning networks (DialogueCRN) to sufficiently understand both situation-level and speaker-level context. 
DialogueCRN introduces the cognitive phase to extract and integrate emotional clues from context retrieved by the perceptive phase.
In the cognitive phase, we design multi-turn reasoning modules to iteratively perform the intuitive retrieving process and conscious reasoning process, which imitates human unique cognitive thinking. 
Finally, emotional clues that trigger the current emotion are successfully obtained and used for better classification.
Experiments on three benchmark datasets have proved the effectiveness and superiority of the proposed model. 
The case study shows that considering cognitive factors can better understand emotional clues and boost the performance of ERC.






\end{document}